\documentclass[letterpaper, 10 pt, conference]{ieeeconf}  %

\IEEEoverridecommandlockouts                              
\usepackage{graphicx} 
\usepackage{amsmath} 
\usepackage{amssymb}  
\usepackage[usenames, dvipsnames]{color}

\usepackage{lipsum}
\usepackage{color}
\usepackage{cite}

\usepackage{diagbox} 
\usepackage{tabu}
\usepackage{balance}  

\usepackage[ruled,linesnumbered]{algorithm2e}
\usepackage{algpseudocode}
\usepackage{epsfig}

\newcommand{\revision}[1]{\textcolor{black}{#1}} 

\title{
Motion-Aware Robotic 3D Ultrasound
}

\author{Zhongliang Jiang*$^{1}$, Hanyu Wang*$^{1}$, Zhenyu Li$^{1}$, Matthias Grimm$^{1}$, Mingchuan Zhou$^{1,2}$, \\Ulrich Eck$^{1}$, Sandra V. Brecht$^{3}$, Tim C. Lueth$^{3}$, Thomas Wendler$^{1}$, and Nassir Navab$^{1,4}$ 
\thanks{$^{*}$ Authors are with equal contributions.}
\thanks{$^{1}$Z. Jiang, H. Wang, Z. Li, M. Grimm, M. Zhou, U. Eck, T. Wendler and N. Navab are with the Chair for Computer Aided Medical Procedures and Augmented Reality, Technical University of Munich, Germany. {\tt\footnotesize{(zl.jiang@tum.de)}}
        }%
\thanks{$^{2}$M. Zhou (Corresponding author) is with the College of Biosystems of Engineering and Food Science, Zhejiang University, China.} 
\thanks{$^{3}$S. V. Brecht and T. C. Lueth are with the Institute of Micro Technology and Medical Device Technology, Technical University of Munich, Germany.}
\thanks{$^{4}$N. Navab is with the Laboratory for Computational Sensing and Robotics, Johns Hopkins University, Baltimore, MD, USA.}
}

\begin{document}

\maketitle

\thispagestyle{empty}    
\pagestyle{empty}

\begin{abstract}
Robotic three-dimensional (3D) ultrasound (US) imaging has been employed to overcome the drawbacks of traditional US examinations, such as high inter-operator variability and lack of repeatability. However, object movement remains a challenge as unexpected motion decreases the quality of the 3D compounding.
Furthermore, attempted adjustment of objects, e.g., adjusting limbs to display the entire limb artery tree, is not allowed for conventional robotic US systems.
To address this challenge, we propose a vision-based robotic US system that can monitor the object's motion and automatically update the sweep trajectory to provide 3D compounded images of the target anatomy seamlessly. To achieve these functions, a depth camera is employed to extract the manually planned sweep trajectory after which the normal direction of the object is estimated using the extracted 3D trajectory. Subsequently, to monitor the movement and further compensate for this motion to accurately follow the trajectory, the position of firmly attached passive markers is tracked in real-time. Finally, a step-wise compounding was performed. The experiments on a gel phantom demonstrate that the system can resume a sweep when the object is not stationary during scanning. 
\end{abstract}





\bstctlcite{IEEEexample:BSTcontrol}
\section{Introduction}
Peripheral artery disease (PAD) is one of the most common diseases, particularly among the elderly, which causes the blocking or narrowing of peripheral blood vessels, thereby limiting blood supply to certain body parts. 
PAD affects $20\%$ of people older than $55$ years, which is estimated to be about $27$ million people in Europe and North America~\cite{hankey2006medical}. In the worst case, PAD causes organ and limb failure, potentially resulting in long-term damage or death.
To diagnose PAD, ultrasound (US) imaging is employed as the primary non-invasive modality in clinical practice~\cite{aboyans20182017}, because it is non-invasive, cheap, and radiation-free, as opposed to other popular imaging modalities such as computed tomography angiography (CTA) and magnetic resonance angiography (MRA). Collins~\emph{et al.} reported that the femoral artery US has $80\%-98\%$ sensitivity in detecting vessel stenoses, even for PAD in its early stages~\cite{collins2007systematic}.

\begin{figure}[ht!]
\centering
\includegraphics[width=0.45\textwidth]{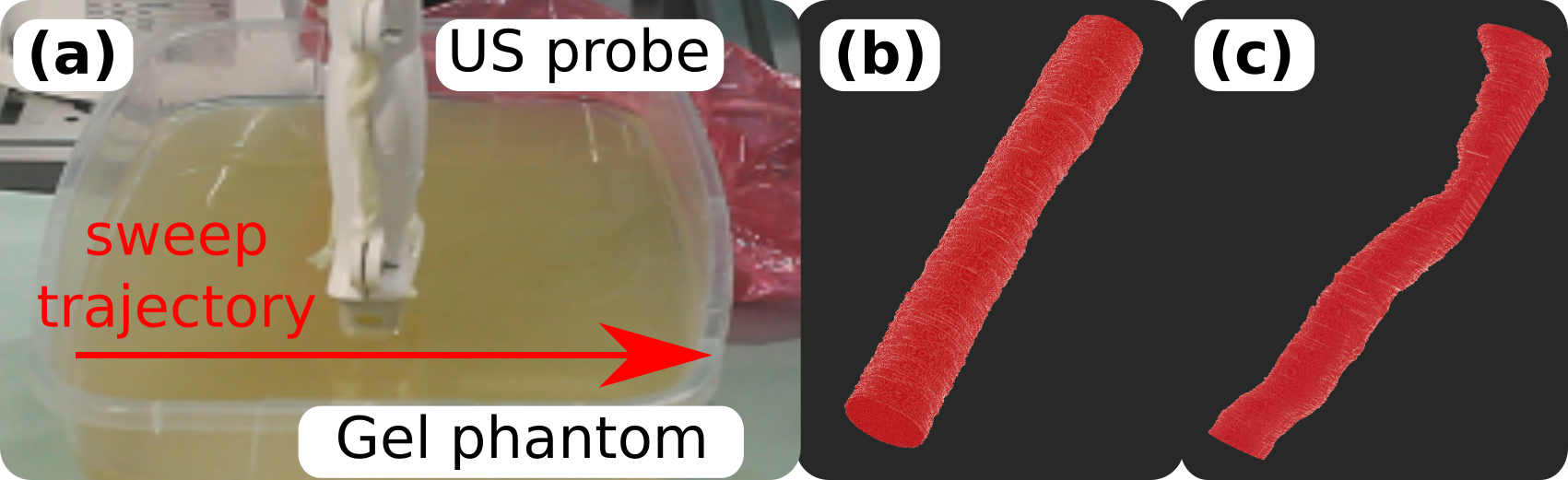}
\caption{Illustration of the influence of object motion on 3D US compounding. (a) US sweep on a phantom containing a straight tube to mimic a blood vessel, (b) 3D reconstruction result of the vessel when the imaged object (phantom) is stationary during the sweep, and (c) 3D US vessel computed when the object is moved randomly relative to the initial trajectory during the sweep. 
}
\label{Fig_shift}
\end{figure}

\par
Traditional two-dimensional (2D) US imaging can show tissues of interest in real-time. However, the lack of 3D anatomy information makes the result sensitive to intra- and inter-operator variability due to the changes in probe orientation and applied force. Therefore, it is challenging to obtain a reliable and consistent PAD diagnosis from different physicians or even from the same physician in different trials. To address this challenge, 3D US was employed to characterize and quantify stenoses non-invasively for carotid~\cite{fenster20063d} and lower limb arteries~\cite{merouche2015robotic}. To augment the 2D slices into the 3D volume, an optical or electromagnetic (EM) tracking system was typically used to record the spatial probe pose (position and orientation)~\cite{gee2003engineering}. 
To overcome the limitation of EM signal interference, Prevost~\emph{et al.} employed an inertial measurement unit to compute free-hand 3D US volumes from long US sweeps~\cite{prevost20183d}. 
The 3D artery geometry is important to determine the extent of PAD and to plan surgery or endovascular intervention~\cite{merouche2015robotic, guo2019novel}. However, the free-hand US acquisition method is inherently limited by non-homogeneous deformations caused by varying contact force between probe and object~\cite{gilbertson2015force}.

\par
To overcome inter-operator variability, robotic US systems (RUSS) are used to produce high-quality US images by accurately controlling the US acquisition parameters (contact force and probe orientation). To avoid non-homogeneous deformation during a sweep, Gilbertson~\emph{et al.} and Conti~\emph{et al.} developed compliant controllers to maintain a constant contact force between the probe and the imaged object~\cite{gilbertson2015force,conti2014interface}. Besides the contact force, Huang~\emph{et al.} employed a depth camera to adjust the probe orientation~\cite{huang2018robotic}. Jiang~\emph{et al.} optimized the probe orientation using force and real-time US images to achieve high image quality~\cite{jiang2020automatic}. Also, to accurately and repeatably position the probe in the normal direction of the object's surface, a model-based method was developed in~\cite{jiang2020automaticTIE}. Since the probe poses can be computed using the known kinematic model, RUSS was also employed to compound 3D US volumes. Huang~\emph{et al.} produced 3D US volumes using a linear stage~\cite{huang2018fully}, and subsequently extended the system to a 6-DOF (degrees of freedom) robotic arm~\cite{huang2018robotic}. Virga~\emph{et al.} presented a framework for acquiring 3D US scans of the abdominal artery~\cite{virga2016automatic}. Since the scan path was generated from a preoperative MR image, the system could not guarantee good performance if the imaged object was moved after calibration. Thus, this method is not suitable for long US sweeps. For example, visualization of the leg artery tree requires motion (potentially rotation) of the leg during acquisition, due to the leg (femoral) artery starting at the inner side of the thigh and spiraling to the knee joint. In summary, previous methods can not handle the motion of the imaged object during a sweep because the object movement (expected and unexpected) has not been taken into consideration. An example of the influence of object motion is shown in~Fig.~\ref{Fig_shift}.

\par

In this study, we present a vision-based, semi-automatic RUSS to generate complete and accurate 3D US volumes of non-stationary imaged objects during US sweeps. To achieve this, a depth camera is employed to monitor potential object motions and provide the information to seamlessly stitch different parts of the sweep together in case of object motion during scans. To the best of our knowledge, this method is a novel approach to automatically compute an accurate 3D US volume for potentially long and complicated tissues such as limb arteries, in the presence of motion.
The method combines the advantages of free-hand US (flexibility) and RUSS (accuracy and stability). The main contributions of this work are summarized as follows:
\begin{itemize}
  \item An adaptive method is proposed to robustly extract planned sweep trajectories drawn on the object's surface and automatically compute key points of trajectories to optimize probe poses for automatic US sweep.
  \item A vision-based movement-monitoring system is proposed to monitor the potential movement of the imaged object and automatically resume the sweep at breakpoints to provide a complete and accurate 3D US volume using a stepwise compounding.
\end{itemize}
Finally, the performance of the proposed motion-aware RUSS is validated on a custom-designed vascular phantom.
\section{Methods}
This section describes the individual components that enable the vision-based RUSS to perform US sweeps for 3D reconstruction robust to object motion. The overall workflow is described in Section II-A, while the RUSS calibration procedures and the method for robustly extracting the planned sweep trajectory are presented in Sections II-B and II-C. Based on the extracted trajectory, the probe pose is optimized in Section II-D, and finally, the movement-monitoring system is described in Section II-E.

\subsection{Vision-Based 3D US System}
\par
The visualization of the limb arterial tree is important for determining the extent of PAD~\cite{merouche2015robotic}. However, vascular structures are longer than other organs, such as the liver. Due to the limitation of the robotic workspace, conventional RUSS cannot handle the case when the desired trajectory is partly out of the robotic workspace. To address this challenge, we propose a RUSS, which allows adjustment of the object position and orientation during the sweep to completely display the anatomy with long geometry. The overall workflow is shown in Fig.~\ref{Fig_workflow}.


\par
\subsubsection{Hardware}
The RUSS comprises a robotic manipulator (LBR iiwa 14 R820, KUKA GmbH, Germany) and a Cicada LX US machine (Cephasonics, USA). A CPLA12875 linear probe (Cephasonics, USA) is attached to the end-effector of the robotic arm and B-mode US images ($50$ fps) are obtained via a USB interface. The robot is controlled via the iiwa stack developed based on ROS~\cite{hennersperger2016towards}.

\subsubsection{Workflow}
The software system consists of three parts: 1) a vision-based sweep trajectory extraction, 2) an automatic robotic US sweep and 3D US compounding, and 3) a movement-monitoring system, which updates the trajectory and corrects the compounding. To realize the system, an RGB-D camera (Azure Kinect, Microsoft Corporation, USA) and passive markers (NDI, Ontario, Canada) are used to obtain three inputs (an RGB image, a depth image, and the position of the markers). Before performing the sweep, the desired trajectory optimizing the visibility of the target anatomy is manually drawn on the patient's skin by medical experts. The drawing is done using a red line, which shows good contrast to the skin color. Afterward, the sweep extraction module is used to extract the drawn trajectory and transform it into the robotic base frame using the hand-eye calibration results. To avoid non-homogeneous deformation and guarantee the patient's safety, a compliant controller maintains a constant contact force in the probe centerline~\cite{jiang2020automaticTIE}. During scanning, a marker-based motion monitor system is activated so that the system can automatically compute the transformation and update the trajectory using the iterative closest point (ICP) method if a motion happens. This correction enables a stepwise compounding.

\begin{figure*}[ht!]
\centering
\includegraphics[width=0.90\textwidth]{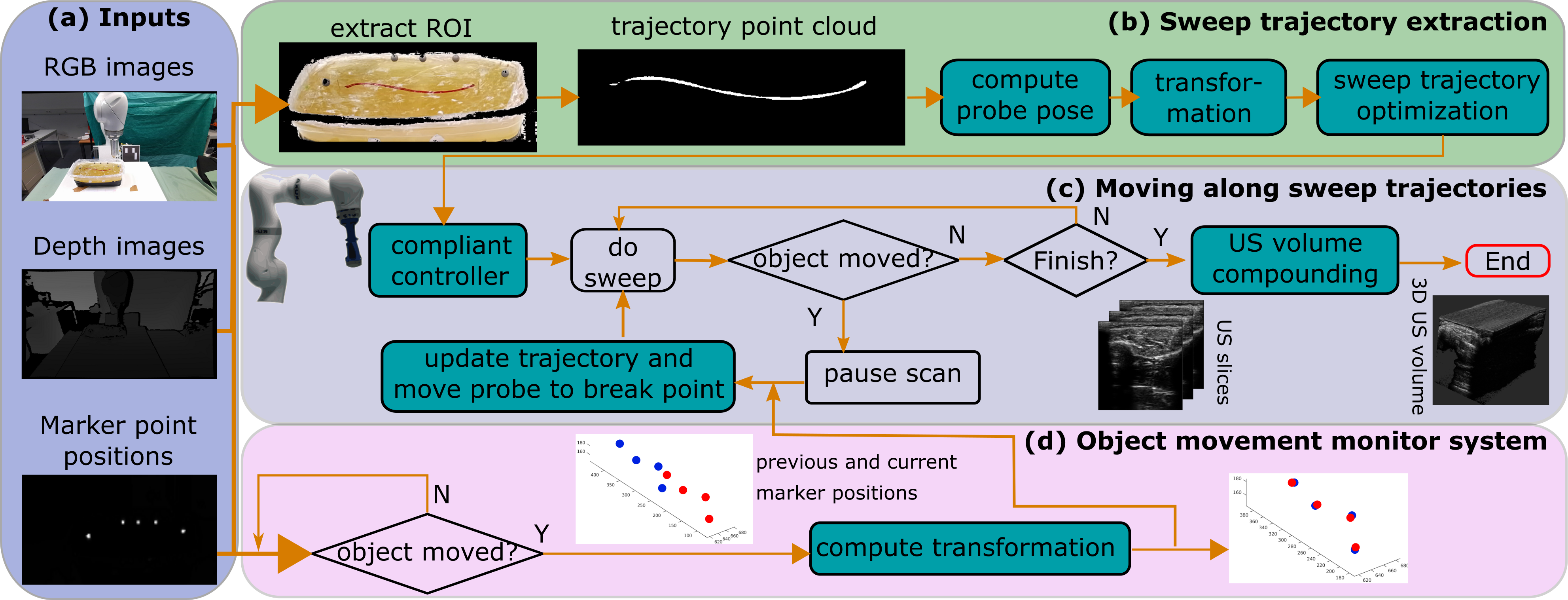}
\caption{System workflow. (a) Three types of inputs from the depth camera, (b) sweep trajectory extraction module, (c) robotic movement module following a planned trajectory, and (d) object movement monitor module.}
\label{Fig_workflow}
\end{figure*}

\subsection{Hand-Eye Calibration}

\begin{figure}[ht!]
\centering
\includegraphics[width=0.45\textwidth]{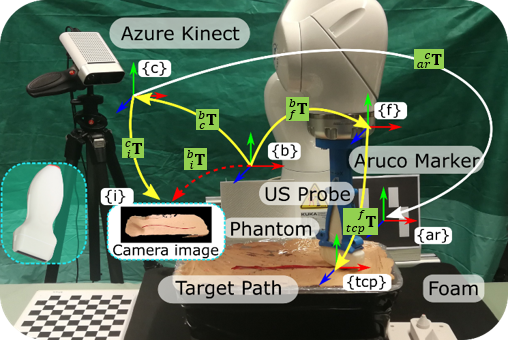}
\caption{Diagram of the involved coordinate frames.}
\label{Fig_coordinate_system}
\end{figure}

\par
To control the robot to follow the manually drawn trajectory, the transformation matrix between the RGB-D camera and the robot was calculated. As shown in Fig.~\ref{Fig_coordinate_system}, the involved coordinate frames are: 1) the image frame \{i\}; 2) the RGB-D camera frame \{c\}; 3) the robotic base frame \{b\}; 4) the robotic flange frame \{f\}; 5) the tool center point \{tcp\}; 6) an ArUco marker~\cite{garrido2016generation} \{ar\}. The transformation from the flange to the base frame $^{b}_{f}\textbf{T}$ can be directly obtained using the robotic kinematic model. Thus, the transformation $^{f}_{tcp}\textbf{T}$ is obtained using the 3D model of the custom-designed probe holder (connecting the US probe to the robot). Besides, the transformation $^{c}_{i}\textbf{T}$, used to generate a 3D point cloud from the 2D image can be computed based on the camera intrinsics accessed via a program\footnote{https://github.com/microsoft/Azure\_Kinect\_ROS\_Driver}. 

\par
The transformation between frame \{b\} and \{c\} $^{b}_{c}\textbf{T}$ can be optimized based on the paired descriptions of the points in different frames using ICP~\cite{sun2018robot}. To accurately estimate $^{b}_{c}\textbf{T}$, at least four non-coplanar points should be employed. Thus, eight arbitrary intersections on two chessboards at different heights are selected. For each point, the position in the camera frame $^{c}P$ is computed using OpenCV and the coordinate expression in frame \{b\} $^{b}P$ is obtained by manually moving the robotic arm to the target intersection. To accurately locate the intersections, a tip-pointed tool, was attached to the flange. Based on the paired position descriptions $(^{c}P$ and $^{b}P)$, the transformation $^{b}_{c}\textbf{T}$ is obtained by optimizing the following equation:

\begin{equation} \label{eq_ICP}
\min_{^{b}_{c}\textbf{T}} \frac{1}{N}\sum_{i=1}^{N}{||^{b}_{c}\textbf{T}~^{c}P_i - ^{b}P_i||^2}
\end{equation}

\par
\revision{To make the calibration system robust to the camera movement, an ArUco marker is fixed relatively to the robotic base as shown in Fig.~\ref{Fig_coordinate_system}. Once $^{b}_{c}\textbf{T}$ is calculated using Eq.~(\ref{eq_ICP}), the fixed $^{b}_{ar}\textbf{T}$ can be calculated using $^{b}_{ar}\textbf{T} = ^{b}_{c}\textbf{T} \cdot ^{c}_{ar}\textbf{T}$. Then, based on real-time $^{c}_{ar}\textbf{T}$, obtained using aruco\_ros\footnote{https://github.com/pal-robotics/aruco\_ros}, $^{b}_{c}\textbf{T}$ can be dynamically updated when the camera is moved.}

\subsection{Extraction of Scan Trajectory}
\par
In this section, we describe the adaptive color-based method for extracting the manually planned trajectory on object's surface. To provide reliable information on the object's position, two passive markers are stuck on the skin (using tape) in both ends of the trajectory as Fig.~\ref{Fig_workflow}.
The marker spheres are covered by a retro-reflective layer, making the markers brighter than the background in the infrared images. Also, the markers aid in robust extraction of the region of interest (ROI) and trajectory (whole trajectory or at least partial trajectory if it is partly occluded).

\par
A multi-color space threshold method was used to exact the ROI~\cite{rahmat2016skin}. The ROI is extracted based on the color feature in the area surrounding the two markers at the end of the trajectory in the Cb and Cr channel images. The upper and lower limits of the ROI are automatically determined based on the value of pixels distributed on the line connecting the two markers. 
Compared with RGB images, YCrCb images separate brightness and chromaticity into different channels. Thus, the negative effect caused by the environment (e.g. light) could be mitigated using the YCrCb color space. To further extract the red trajectory from the ROI, an adaptive threshold method on both the Cr channel image and the grayscale image is proposed. To robustly extract the trajectory line, even when partly blocked by the probe, $N_s$ seeds lines are initialized to equally divide the space between the two markers at the end of the trajectory. Then, the intersection points $P_s$ between the seed line and the trajectory can be detected by locating the maximum Cr value of the pixels on the seed line (Fig.~\ref{Fig_key_point_detection}). 

\par
However, since the probe may partly block the trajectory in 2D images, the detected points $P_s$ may not be on the trajectory. To further remove these points from $P_s$, the ``up'' \revision{($Y_p$)} and ``down'' \revision{(-$Y_p$)} boundaries of the trajectory for each $P_s$ in the Cr channel images can be calculated using:

\begin{equation} \label{eq_boundaries}
\begin{array}{lr}
    f_b^{up} = max\left( \left[|I(x_i, y_{i}+j+1)-I(x_i, y_{i}+j)| \right]\right)  \\
    f_b^{down} = max\left( \left[|I(x_i, y_{i}-j-1)-I(x_i, y_{i}-j)| \right]\right)
\end{array}
\end{equation}
where $(x_i, y_i)$ represents the $i$-th $P_s$, $j$ is $1,2,3,...,8$. 

If the boundary features $f_b^{up}$ and $f_b^{down}$ are close to zero ($<10$), the boundary features on the grayscale images are computed again using Eq.~(\ref{eq_boundaries}). If the features for $P_s(i)$ computed based on the grayscale images are still close to zero, it will be removed from $P_s$ because the intensity of the real $P_s$ are supposed to be significantly different from the skin background.

\par
Based on the computed seed points $P_s$ located on different parts of the trajectory, a bidirectional searching method starting from each $P_s$ is developed. Considering the potential color differences between different trajectory parts, an adaptive threshold is used to effectively extract the trajectory. First, a moving box ($B_w\times B_h$) is initialized at $P_s(i)$. Then, the number of points, whose intensity is between $[I(P_s))$-$I_rt,~I(P_s)+I_rt]$, is counted as $N$. If $N$ is larger than the empirical threshold $N_{ep}$, at least a short part of the trajectory is located inside the current box. To further extract the trajectory, the moving box and the corresponding local threshold range are updated using the last detected trajectory point. These procedures have been described in lines $5-18$ in Algorithm~\ref{algorithm_extraction_path}. Like these, another moving box is initialized at $P_s(i)$ to search in the inverse direction. The seeds-based bidirectional searching modality enables the proposed method to provide the most possible result. 
The involved parameters are empirically set to: $B_w = 20~pixels$, $B_h = 100~pixels$, $I_{rt} = 25$, and $N_{ep}=10$.

\begin{algorithm}[htb] 
\caption{Adaptive Trajectory Extraction}\label{algorithm_extraction_path}
\KwIn{seed points set $P_s = (x_s, y_s)$, moving box width $B_w$, moving box height $B_h$, relative threshold of Cr channel $I_{rt}$ and the number of the pixels extracted in the moving box $N_{ep}$}
\KwOut{2D trajectory $P_t^{2d}$}
\For{$i=2$; $i<\text{len}(P_s)$; $i++$}
{
    \textbf{Searching from $P_s(i)$ to $P_s(i+1)$}\;
    \For{$x=x_s(i)$; $x<x_s(i+1)$}
    {
        $N ~\xleftarrow~ 0$ \;
        \For{$m = x$; $m<x+B_w$; $m++$}
        {
            \For{$n=y_s(i)-\frac{1}{2}B_h$; $n<y_s(i)+\frac{1}{2}B_h$; $n++$}
            {
                \If{$I(m,n)\in~[I(x, y_s(i))-I_rt,~I(x, y_s(i))+I_rt]$}
                {
                    $P_t^{2d}~\xleftarrow~[P_t^{2d}, (m,n)]$ \;
                    $N~\xleftarrow~N+1$ \;
                }
            }
        }
        \If{$N>N_{ep}$}
        {
            $(x, y_s(i))~\xleftarrow~(P_t^{2d}(end))$\;
        }
        \Else{
            Break;
        }
    }
    \textbf{Similar to 3-19, searching from $P_s(i)$ to $P_s(i-1)$;}
}
\end{algorithm}

\subsection{Probe Orientation Determination and Optimization}
\par
Based on the 2D trajectory $P_t^{2d}$ extracted in the last section, the 3D trajectory point cloud $P_t^{3d}$ is computed using the camera internal parameters and used to control the position of the probe. However, $P_t^{3d}$ only tells the desired positions of probe during the scan. To receive more signal back to the US elements built in US probe tip, Jiang~\emph{et al.} and Ihnatsenka~\emph{et al.} suggested that the probe should be aligned in the normal direction of the contact surface $\Vec{n}_i$~\cite{jiang2020automatic, ihnatsenka2010ultrasound}. Here, a vision-based normal direction estimation method was proposed to quickly compute the desired poses for the whole sweep rather than a force-based method, which only works for the current contact position used in~\cite{jiang2020automatic}. Compared with~\cite{huang2018fully}, more local points distributed around the point on the trajectory $P_t^{3d}$ are considered to accurately and stably estimate $\Vec{n}_i$. The selected point will be located on a plane when the local points are distributed close enough. In that case, $\Vec{n}_i$ at $P_t^{3d}(i)$ is approximated by the normal direction of the plane. The plane expression ($z = \text{f(x, y)}$) is optimized using the Least-Squares method as Eq.~(\ref{eq_fiting}).  

\begin{equation} \label{eq_fiting}
\min_{f(x_i,y_i)} \frac{1}{2N}\sum_{i=1}^{N}{(f(x_i,y_i) - z_i)^2}
\end{equation}


\par
After aligning the probe to the estimated $\Vec{n}_i$, the probe tip is expected to be perpendicular to the scan trajectory. Since the width of the manually drawn trajectory varies, the moving direction computed by connecting two close points in $P_t^{3d}$ may differ significantly from the real value, causing instability in the rotation around the probe centerline (aligned with $\Vec{n}_i$) during scanning. To address this problem, we propose a difference-based optimization method to automatically select key points from $P_t^{3d}$,  generating a smooth robotic movement trajectory. To achieve this, the 3D points $P_t^{3d}$ are transformed into 2D vectors $(x_p(i), y_p(i))$ as follows:

\begin{equation} \label{eq_3DTo2D}
(x_p(i), y_p(i)) = (\overrightarrow{P_{s} P_t^{3d}(i)} \cdot \textbf{X}_p,~|\overrightarrow{P_{s}P_t^{3d}(i)} \times \textbf{X}_p|)
\end{equation}
where $\textbf{P}_s$ is the start point of the path and $\textbf{X}_p$ is a unit vector connecting start and end points $\textbf{P}_e$ as $\textbf{X}_p=~\frac{\textbf{P}_e-\textbf{P}_s}{|\textbf{P}_e-\textbf{P}_s|}$. 

\begin{figure}[ht!]
\centering
\includegraphics[width=0.40\textwidth]{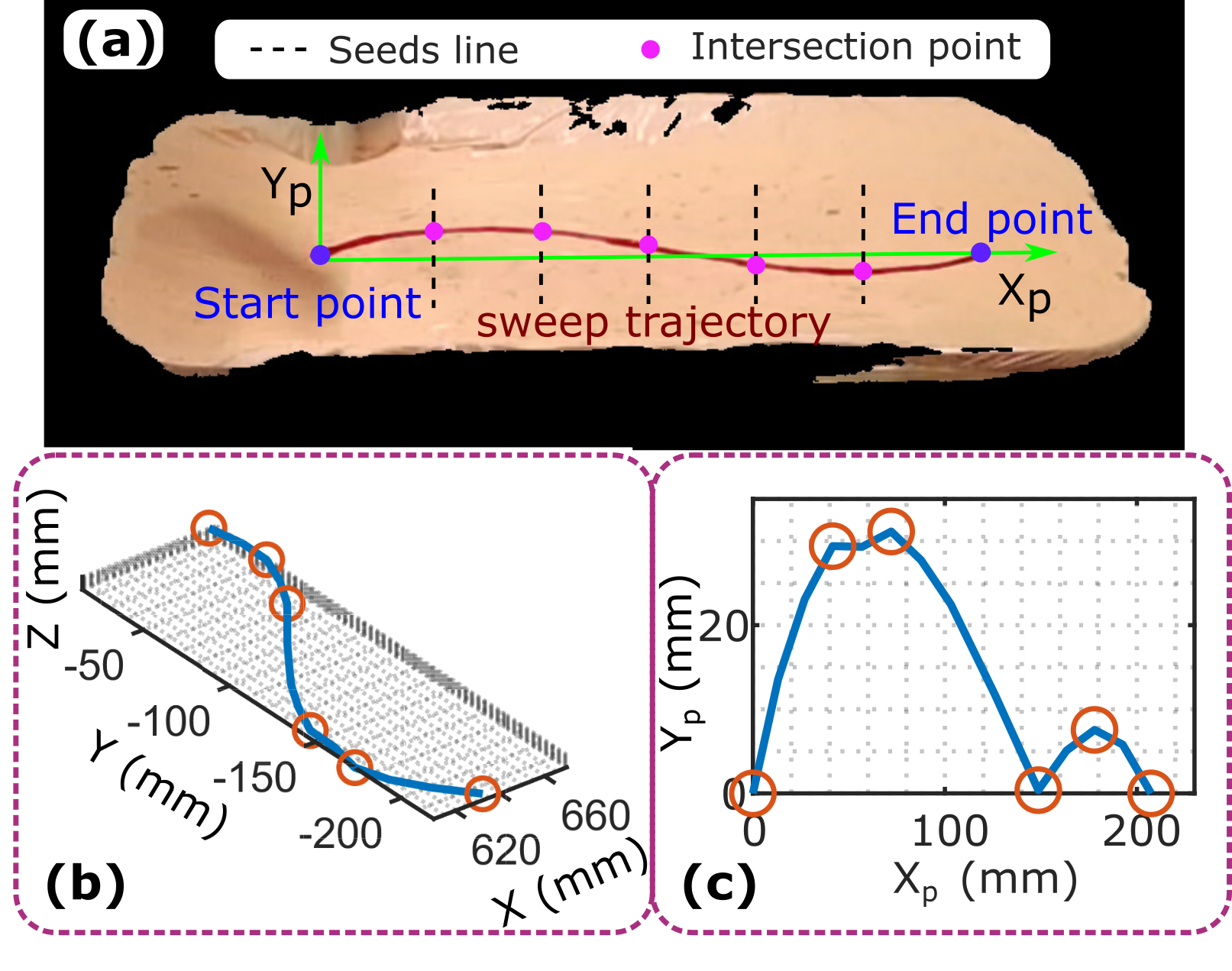}
\caption{Trajectory optimization of the robotic movement. (a) Planned trajectory on the object's surface. (b) and (c) are automatically detected key points in 3D space and 2D space, respectively.}
\label{Fig_key_point_detection}
\end{figure}

Based on the transformed 2D position vector $P_p(x_p, y_p)$, the local maxima and minima are extracted as follows:
\begin{equation} \label{eq_first_order_diff}
P_k(i)~~i\in\{k|D(k-1)D(k)\leq0\&|D(k-1)|+|D(k)|>T_k\}
\end{equation}
where $D(i) = y_p(i+1) - y_p(i)$ is the first-order difference and $T_k$ is the threshold used to remove the local extrema with a small amplitude. Fig.~\ref{Fig_key_point_detection} shows an illustration of the key points in both 2D and 3D, where all turning points have been correctly detected as key points (marked as red circles).

To restrain the effect caused by the local identification error of estimated probe orientation, the orientation is further optimized for each interval separated by the key points and the start and stop points.
To improve movement stability, points located close to their neighbors are not used to define an interval as lines $5-7$. For each interval, the desired direction of the probe centerline ($^{tcp}\textbf{Z}$) is represented by the mean $\Vec{n}_i$ computed at all sampled positions. Afterward, the probe's long side direction ($^{tcp}\textbf{Y}$) is placed in the normal direction of the plane consisted by $^{tcp}\textbf{Z}$ and the line connecting the two $\textbf{P}_k$ defines the interval. Finally, the moving direction ($^{tcp}\textbf{X}$) is perpendicular to $^{tcp}\textbf{Z}$ and $^{tcp}\textbf{Y}$.
The implementation details are described in Algorithm~\ref{algorithm_Key_points}. 

\begin{algorithm}[htb] 
\caption{Key points Based Orientation Optimization}\label{algorithm_Key_points}
\KwIn{Planned path $\textbf{P}_t^{3d}$, start point $\textbf{P}_s$, end point $\textbf{P}_e$, Key points on path $\textbf{P}_k$, normal direction of tissue $\Vec{n}_i$ at each $\textbf{P}_t^{3d}(i)$, and threshold $T_1$} 
\KwOut{Segmented intervals $\textbf{Segs}$, optimized probe orientation of all intervals $[^{tcp}\textbf{X}, ^{tcp}\textbf{Y}, ^{tcp}\textbf{Z}]$}
Add $\textbf{P}_s$ and $\textbf{P}_e$ to $\textbf{P}_k$: $\textbf{P}_k \xleftarrow{} [\textbf{P}_s, \textbf{P}_k, \textbf{P}_e]$ \;
Compute iteration set of $\textbf{P}_k$ in $\textbf{P}_t^{3d}$: $\textbf{I}=find(\textbf{P}_k)$\;
int $m=1$\;
\For{$j=1; j < \text{len}(\textbf{P}_k)$} 
{
    \If{$|\textbf{I}(j+m)-\textbf{I}(j)|*|\textbf{P}_k(j+m)-\textbf{P}_k(j)|\le T_1$}
{
   $m \xleftarrow~m+1$\;
} 
    \Else{
    $\textbf{Segs} \xleftarrow~[\textbf{Segs}, [I(j), I(j+m)]]$\;
    $^{tcp}\textbf{Z}\xleftarrow~[^{tcp}\textbf{Z}, \frac{1}{I(j+m)-I(j)}\sum_{i = I(j)}^{I(j+m-1)}\Vec{n}_i]$\;
    $^{tcp}\textbf{Y}\xleftarrow~[^{tcp}\textbf{Y}, \frac{1}{I(j+m)-I(j)}\sum_{i = I(j)}^{I(j+m-1)}\Vec{n}_i \times \overrightarrow{\textbf{P}_t^{3d}(I(j))\textbf{P}_t^{3d}(I(j+m))}]$\;
    $ \textbf{Pose}= [\textbf{Pose}, [^{tcp}\textbf{Y}\times ^{tcp}\textbf{Z}, ^{tcp}\textbf{Y}, ^{tcp}\textbf{Z}]]$\;
    $j \xleftarrow~j+m$\;
    $m \xleftarrow~1$\;
    }
}
$[^{tcp}\textbf{X}, ^{tcp}\textbf{Y}, ^{tcp}\textbf{Z}] \xleftarrow~ \textbf{Pose}$\;
\end{algorithm}

\subsection{Movement Monitor System}
\par
To monitor potential object motions, the markers attached to the ends of the trajectory are used as inputs to the monitor system. The robot immediately stops when the system detects a change in the position of marker over a given threshold ($5~mm$). 
To accurately update the trajectory to resume the sweep from the break point, three additional markers are randomly stitched on the object. Next, the transformation between the previous and current object poses is calculated using ICP based on four paired markers positions as shown in Fig.~\ref{Fig_workflow}~(d). The last marker is used to compute the error of the ICP results as follows: 

\begin{equation}\label{eq_movement_compesation_error}
e_{mc}=\left \|\mathbf{P}_m^{'}-(\mathbf{R}_{mc} \cdot \mathbf{P}_m + \mathbf{T}_{mc})\right\|
\end{equation}
where $\mathbf{P}_m$ and $\mathbf{P}_m^{'}$ are the positions of the passive markers before and after the movement, respectively, while $\mathbf{R}_{mc}$ and $\mathbf{T}_{mc}$ are the computed rotation matrix and translation vector.

\par
The RUSS automatically resumes the sweep from the breakpoint only when $e_{mc}$ is small enough ($<1~cm$). Otherwise, the RUSS automatically ends the sweep.

\section{Results}

\subsection{Vision-based Trajectory Extraction Results}
\par
\begin{figure}[ht!]
\centering
\includegraphics[width=0.40\textwidth]{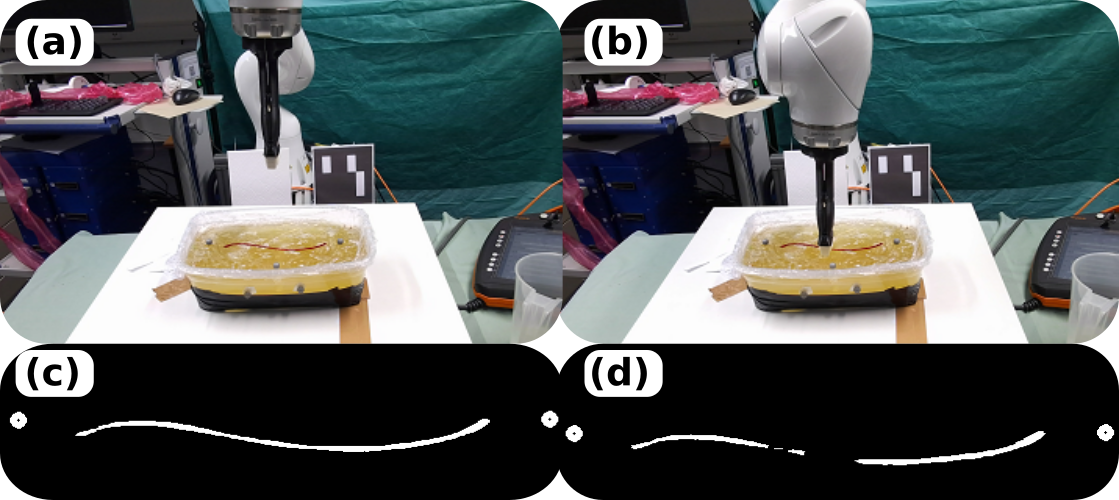}
\caption{Results of the trajectory extraction method. RGB images obtained from the depth camera when the trajectory can be (a) completely or (b) partially seen, depending on the probe position. (c) and (d) are the detected results of (a) and (b), respectively. The line represents the detected trajectory, while the white circles represent the passive markers.}
\label{Fig_path_detection}
\end{figure}

To validate the performance of the adaptive trajectory extraction method described in Section II-C, a gel phantom with a manually drawn trajectory on its upper surface was employed. The phantom was randomly placed in different places inside the camera view. To show the result, an RGB image with the detected trajectory is presented in Fig.~\ref{Fig_path_detection}~(a) and (c), respectively. Besides, considering that the trajectory will be partly obstructed by the robotic arm and the US probe during the sweep, the probe on the top of the trajectory was further moved as in Fig.~\ref{Fig_path_detection}~(b). The proposed adaptive method completely shows the unblocked part of the trajectory as Fig.~\ref{Fig_path_detection}~(d). The detected results are close to the trajectory in Fig.~\ref{Fig_key_point_detection}~(a). More examples of the detected results can be found in this video\footnote{https://https://www.youtube.com/watch?v=8IaorIl3zzk}.

\par
\begin{figure}[ht!]
\centering
\includegraphics[width=0.45\textwidth]{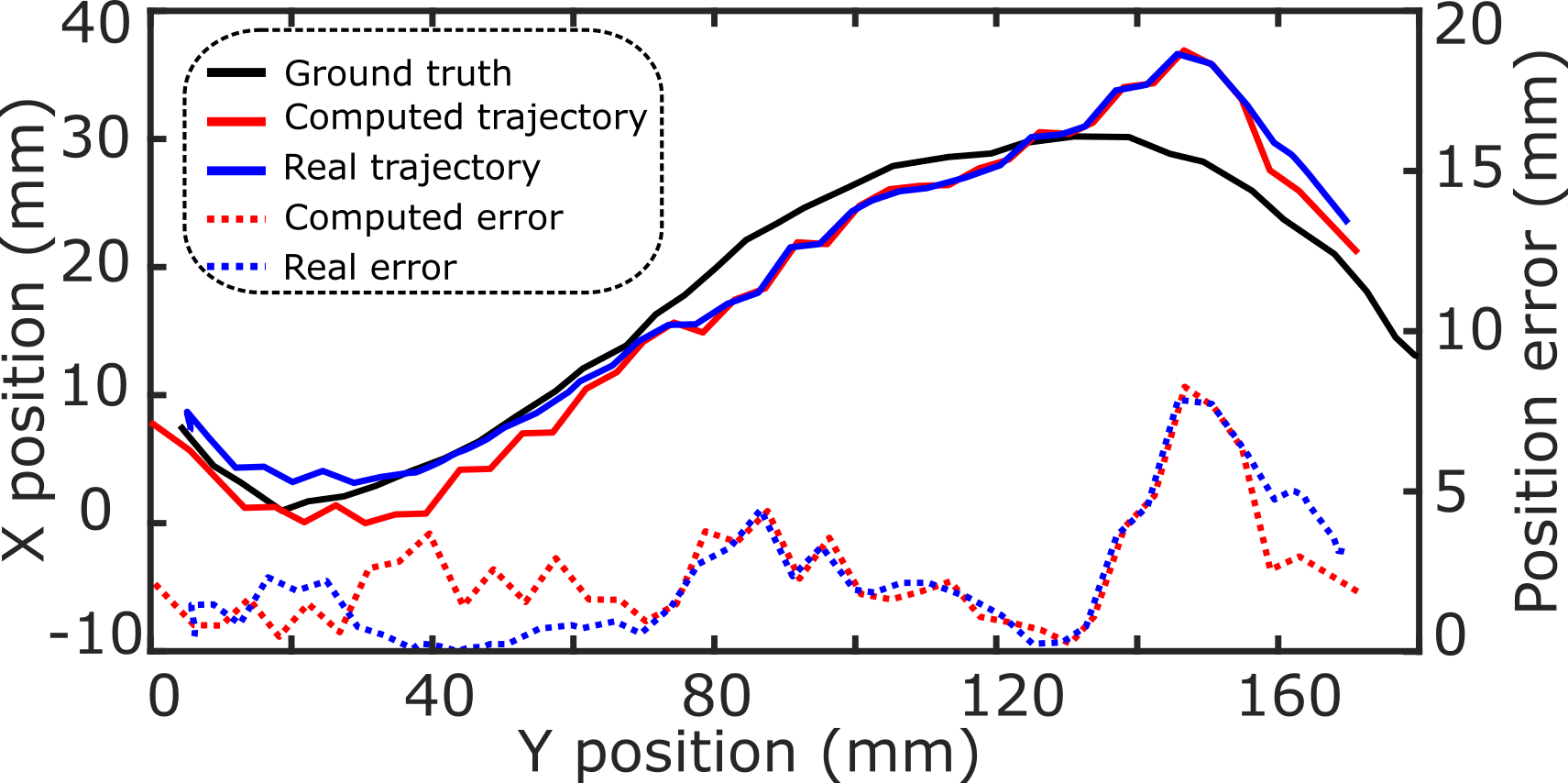}
\caption{Trajectory following results. The solid lines represent the trajectories. The dotted lines are the computed and real position error.}
\label{Fig_2DPosition_error}
\end{figure}

\revision{
Since the direction of the probe centerline ($^{tcp}\textbf{Z}$) is controlled in the force mode~\cite{jiang2020automaticTIE}, the probe tip along the trajectory in the  $^{tcp}X$-$^{tcp}Y$ plane should be accurately moved. To quantitatively analyze the whole system accuracy (trajectory detection and hand-eye calibration), the computed trajectory was transformed into the robotic frame using the result of hand-eye calibration. To demonstrate the real position error between the desired trajectory and the performed path, the position of the probe tip was recorded. Besides, the computed error is calculated based on the computed trajectory and ground truth. The ground truth is obtained by manually guiding the robot along the manually drawn trajectory. The ground truth, the computed trajectory, and real trajectory are depicted in Fig.~\ref{Fig_2DPosition_error}. }


\par
As shown in Fig.~\ref{Fig_2DPosition_error}, both the computed and real error mainly distribute below $5~mm$. Also, the error is not cumulative.
The error still can be close to zero after a large error, for example when $Y=150~mm$ (Fig.~\ref{Fig_2DPosition_error}). Generally, the average computed and average real errors ($\pm SD$) were $2.5\pm1.8~mm$ and $2.0\pm1.9~mm$, respectively. The computed error was mainly caused by the depth estimation. Since our application must accurately control the probe in the $^{tcp}X$-$^{tcp}Y$ plane, the accuracy of the system can be further improved by fixing the camera on the top of the target objects to reduce the negative influence caused by inaccurate depth estimations.


\subsection{Movement Compensation Results}
\par
To validate the performance of the movement compensation algorithm, experiments were carried out on a gel phantom. To further investigate the method's sensitivity to the movement types (translation and rotation), the experiments were grouped into two sets. For the translation set, the phantom was moved along a straight line at different distances ($50, 100, 150,~and~ 200~mm$). For the rotation set, the phantom was rotated around a fixed axis at different angles ($10, 20, 30,~and~40^{\circ}$). Each set consisted of $10$ independent experiments. \revision{The translated and the rotated axes are randomly set in the camera view.}

\par
\begin{figure}[ht!]
\centering
\includegraphics[width=0.37\textwidth]{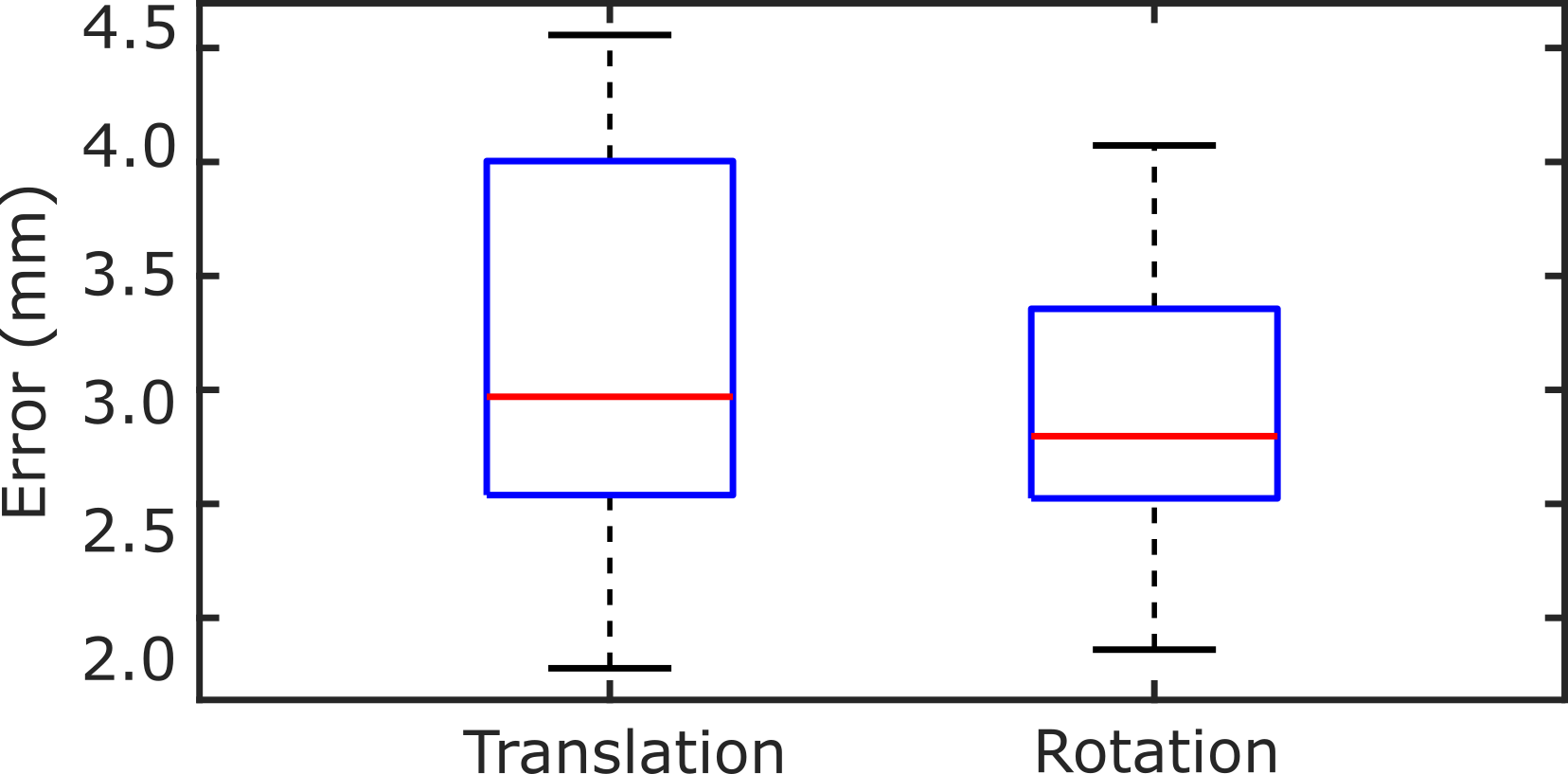}
\caption{Absolute error of the movement compensation algorithm. Two motion types (translation and rotation) over $10$ trials are displayed.}
\label{Fig_compensation_error}
\end{figure}

To assess the performance of the marker-based compensation method, the errors $e_{mc}$ (Eq.~(\ref{eq_movement_compesation_error})) of the two groups of experiments are presented in Fig.~\ref{Fig_compensation_error}. 
The absolute translation and absolute rotation errors ($\pm$SD) were $3.1\pm1.0~mm$ and $2.9\pm0.7~mm$, respectively. Also, based on a t-test (probability $p=0.67>0.05$), it was concluded that no significant difference exists between the two experimental sets. Furthermore, the results also show that most errors ($>75\%$) are less than $4.0~mm$ even when the phantom was moved $200~mm$ and rotated $40^{\circ}$. Such $e_{mc}$ is promising in clearly displaying the partitioned sweeps, which generate complete geometry of the anatomy after a larger motion because $e_{mc} = 4~mm$ is much smaller than the probe width $37.5~mm$.

\begin{figure}[ht!]
\centering
\includegraphics[width=0.4\textwidth]{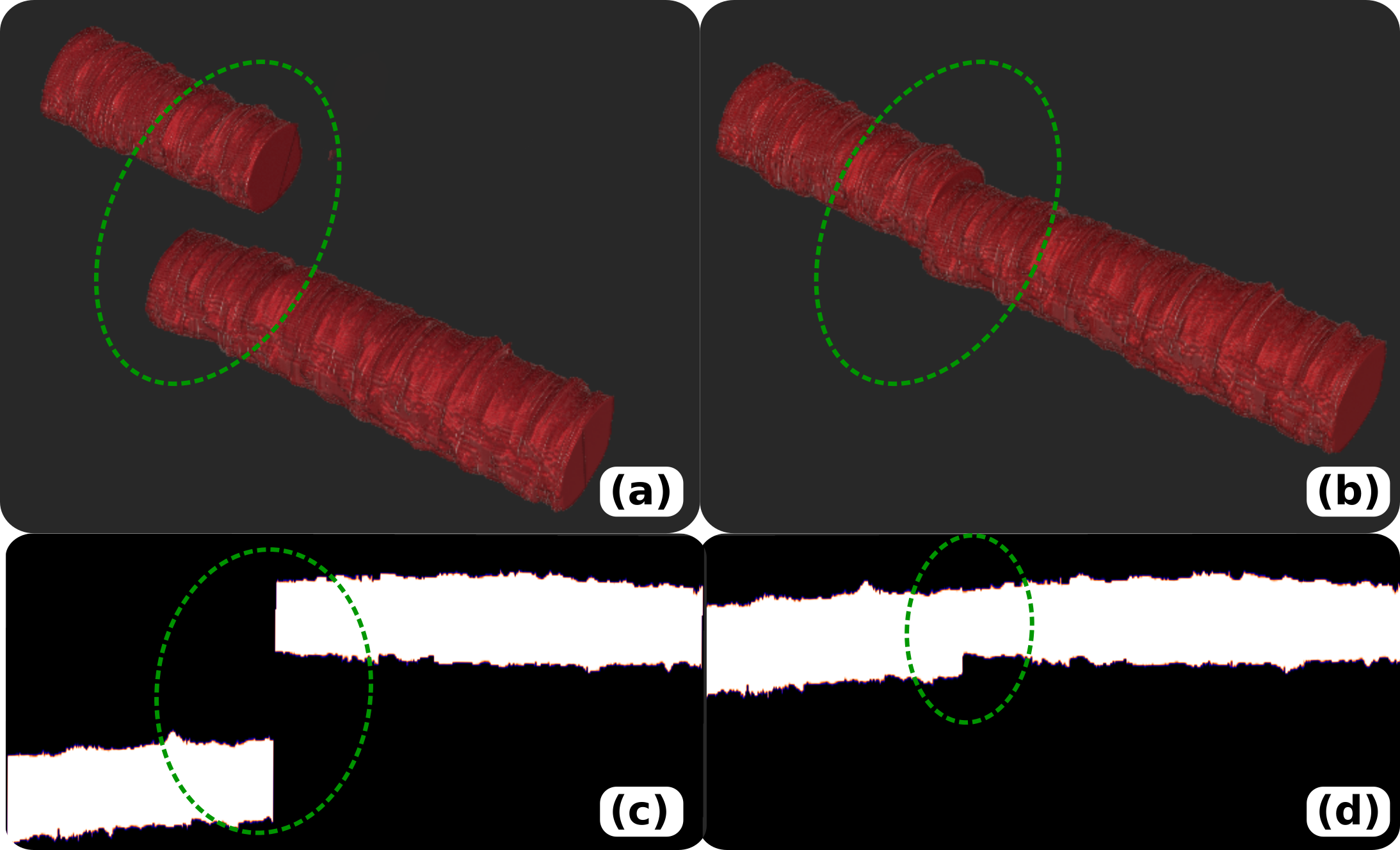}
\caption{Performance of compensation method. 3D images of a straight vessel (a) without and (b) with motion compensation. (c) and (d) are the corresponding 2D images (axial plane) of (a) and (b), respectively.}
\label{Fig_3D_compesated}
\end{figure}

To intuitively show the performance of the compensation method on 3D compounded results, experiments were carried out on a gel phantom with a straight hole, mimicking the blood vessel, which was automatically segmented from the cross-sectional US images using a well-trained U-Net from~\cite{jiang2020autonomous}. Then, the 3D compounding process was performed using ImFusion Suite (ImFusion GmbH, Munich, Germany). The 3D vessels with and without motion compensation are shown in Fig.~\ref{Fig_3D_compesated}. The result reveals that the proposed approach can deliver a complete 3D image of the target blood vessel even
in the presence of object motion. 
\revision{Limited by the accuracy of depth estimation, the stitching is still visible using the current setup. But the stitching error could be reduced using a camera with an accurate depth estimation.}

\section{Discussion}
The preliminary validation on a gel phantom demonstrates that the proposed approach can provide a promising 3D geometry even when the scanned object is moved. In this work, the trajectory was drawn using a marker pen, which can be replaced by a laser system to automatically project a sweep trajectory on the surface. Since the relative threshold was determined based on local seed points, a promising result can be computed if the trajectory color significantly differs from the skin color. However, since the skin color was approximated using the passive markers, the visibility of the markers must be guaranteed during the initialization procedure before executing scan. This affordable system is developed to completely visualize the anatomy with long structures by automatically stitching the US sweeps acquired when the object is moved. Although the vascular application was used to demonstrate the proposed methos, the method can also be used for other applications such as US bone visualization. Regarding the motion compensation, the proposed method so far considered the large object motions during US scans. The small physiological motions ($<5~mm$), such as tremor, vessel motion~\cite{chen20163d} and breathing~\cite{jiang2017model, jiang2020model} have not been considered. Since a compliant controller was used, the probe is flexible in its centerline direction. Thus, compared with free-hand US acquisition, the effect of such small motions is not aggravated in our setup. If we further consider the articulated limbs, more markers could be used to compute the transformation of each rigid parts.

\section{Conclusion}
This work introduces a motion-aware robotic US system. A vision-based system is built for monitoring object motions and automatically update the sweep trajectory to seamlessly accomplish a planned sweep trajectory using a RUSS. This method accurately enables compounding the 3D US volume of a tissue of interest, even when the object is moved during US scanning. In this way, the advantages of both free-hand US (flexibility) and robot US (accuracy and stability) can be integrated into one system. Thus, making the proposed RUSS useful for visualizing tissues with long structure, such as limb artery tree. The results show that the proposed method can accurately compensate a translation movement ($3.1\pm1.0~mm$) and rotation movement ($2.9\pm0.7~mm$) for movements of up to $200~mm$ and rotations of up to $40^{\circ}$. For future development, this approach can be optimized for routine use, in particular, given that it does not require any pre-interventional imaging, but merely the clinical know-how of an expert to define the trajectory on the skin. Such improvements we believe can make RUSSs more robust and thus, bring them closer to clinical use.

\bibliographystyle{IEEEtran}
\balance
\bibliography{IEEEabrv,references}

\end{document}